\documentclass[10pt,conference]{IEEEtran}
\usepackage{url}
\usepackage{graphicx}  
\usepackage{wrapfig}  
\usepackage{multicol}
\usepackage[linesnumbered,ruled,vlined]{algorithm2e}

\newcommand{\eat}[1]{}
\newcommand{\system}{AITEST}

\usepackage{wrapfig}
\usepackage{amssymb}
\usepackage{amsmath}
\usepackage{graphics}
\usepackage{graphicx}
\usepackage{latexsym}
\usepackage{array}
\usepackage{enumitem}
\newcolumntype{M}[1]{>{\centering\arraybackslash}m{#1}}
\usepackage{multirow}
\usepackage{url}
\usepackage{listings,color,colortbl}

\definecolor{light-gray}{gray}{0.80}

\usepackage{balance}

\begin{document}

\title{Testing Framework for Black-box AI Models}

\author{\IEEEauthorblockN{Aniya Aggarwal, Samiulla Shaikh, Sandeep Hans, Swastik Haldar, Rema Ananthanarayanan, Diptikalyan Saha}
\IEEEauthorblockA{IBM Research AI}
{\{aniyaagg,samiullas,shans001,swhalda1,arema,diptsaha\}}@in.ibm.com
}

\maketitle

\begin{abstract}
With widespread adoption of AI models for important decision making, ensuring reliability of such models remains an important challenge. In this paper, we present an end-to-end generic framework for testing AI Models which performs automated test generation for different modalities such as text, tabular, and time-series data and across various properties such as accuracy, fairness, and robustness. Our tool has been used for testing industrial AI models and was very effective to uncover issues present in those models. \\
Demo video link- \url{https://youtu.be/984UCU17YZI} 
\end{abstract}

 
\section{Introduction}
\label{sec:intro}

AI models are increasingly being used in many real-life applications like criminal justice \cite{skeem2007assessment}, credit scoring \cite{west2000neural} and healthcare \cite{gomez2011neural}, \cite{davenport2019potential}. While such models are powerful in solving complex decision making, the trustworthiness of these models remains a big issue for their broad adoption. Due to its effectiveness in finding faults, testing is a popular technique for ensuring the reliability of software systems. However, there has not been a generic/holistic testing framework for testing AI models~\cite{MLTEST} even though such a system is of utmost importance~\cite{FORR}. 

\eat{
Most of the efforts in the literature try to improve various kinds of AI models against one or more of these parameters. But surprisingly, there are very few attempts at developing generic protocols and systems for testing the complex AI models.
There are indeed many general software testing frameworks available \cite{5386909}, \cite{1265525}. In AI testing, there are many custom testers for specific models~\cite{checklist}., each of which operates in isolation for a specific type of model and re-use across different models is not straightforward. A platform which supports re-use of different AI testers across different models would be very useful for AI developers and testers. However, building such a platform specifically for testing multiple kinds of AI models presents many challenges:
}

\noindent{\bf Challenges.} In this paper, we present our AI Testing framework called \system{}. We first enumerate the research and software engineering challenges while clearly stating the choices related to AI Testing that motivated the design of \system{}.

\begin{itemize}[leftmargin=*]
    \item Multiple Modalities: Unlike software systems that mainly work on structured data, AI systems apply to both structured (tabular, time-series) and unstructured (text, image, audio, video) data. The algorithms are often modality-specific because of the variations in encodings, distance functions, and data properties. 
    \item Property/Oracle: AI systems demonstrate various issues related to accuracy, fairness, and  robustness~\cite{MLTEST} and therefore, automated testing needs to cover such varied properties.  However, 
accuracy testing requires gold-standard labels which demands much manual effort to curate them. So, the challenge is to devise important metamorphic properties~\cite{segura2016survey} which obviates the need for a test oracle.
    \item Insufficient Test Data: Current practice only uses a fraction of available data for testing that is often insufficient to uncover faults. Therefore, the challenge is to generate high-coverage realistic test data that is effective at discovering faults.  Further, the challenge is to synthesize data from the user-specified data distribution to perform `what-if' testing. 
    \item Interpretability/Explainability: Most AI models are not interpretable and therefore, explaining a fault remains a huge challenge, unlike software systems where whole/partial traces and data dumps can be shown as failure explanations.   
    \item Access. Depending on which stage of AI pipeline the testing appears, the tester gets white-box or black-box access to the model. The black-box access enables techniques to generalize across various types of models, whereas white-box techniques can be more effective in testing as it has access to more structural information.   
\end{itemize}

Other than the usual capability of test case execution, management, and the right user-interface, building an AI testing framework needs to address the following software engineering challenges.

\begin{itemize}[leftmargin=*]
    \item Flow/Pipeline. The challenge is to define a single end-to-end simple pipeline to test varied types of models and properties. 
    \item Abstraction. Given the variety of properties for AI testing, the definition of the test-case itself is very different. Defining the right abstraction for the underlying data model will make the framework more generic.  
    \item Extensible. Given that new algorithms will keep emerging for AI testing, the framework should expose functionalities to extend the testing capability with minimal effort.  
\end{itemize}

\eat{

\begin{itemize}
\item The framework should be general enough to incorporate multiple testers that will be needed for the various kinds of models, which may have different modalities like text, tabular data, time series data and conversation data among others. The workflow needs to be generic enough to accommodate the requirements of each kind of tester.
\item Each tester is associated with a particular type of model and the framework should provide a common interface to accommodate any kind of tester.
\item One of the known challenges in AI testing is the lack of sufficient amounts of testing data. One approach to solve this has been to develop testers that generate data, either synthesizing new data or perturbing existing data sets. The framework should be able to smoothly inter-operate between  the different input-output formats of the models and the formats in which the tester generates the data.
\item Different testers can report different metrics. The framework should define common specifications for on-boarding testers, each of  which may report execution status, results, and other metrics on each execution.


\end{itemize}
}

Even with these challenges, there are certain common grounds that help us in generalizing the process of testing the AI models. With this background in mind, we have developed a testing framework which automatically generates test cases to test variety of properties like correctness, fairness, sensitivity, robustness for classification models built on tabular or text data, and prediction models on time-series data. Our tool requires only black-box access to models and is therefore model-agnostic. Using our tool, \emph{AI testers} as well as \emph{model developers} can evaluate AI models in a systematic and efficient way, thereby significantly speeding up the end-to-end process of deploying AI models.

In this paper, we describe our framework focusing on software engineering challenges without going deep into the algorithmic aspect of solving the research challenges for which we refer to our previous works~\cite{TR,testing-fse}. Our tool is part of IBM's Ignite quality platform~\cite{IGNITE}.  

\eat{
A recent effort \textcolor{red}{cite our long paper here} \cite{our_long_paper} \footnote{This demo paper is the implementation of the cited paper. The demo includes additional modalities.} that proposes a testing framework for the AI models mainly focuses on structured tabular data, but the underlying idea of generating artificial test data and using it to evaluate the model is generic enough to extend for all the possible modalities of the AI models.
}

The rest of the paper is organised as follows. 
In Section~\ref{sec:system}, we discuss the system design, architecture and the workflow. Section~\ref{sec:algo} briefly describes the functionalities of the underlying algorithms along with a short description of the key results. The related work is briefly mentioned in Section~\ref{sec:related}. We conclude with a summary and future work in Section~\ref{sec:conclusion}.

\section{System}
\label{sec:system}

In this section, we present 1) the flow/pipeline of \system{}, 2) the underlying data model, and 3) the generic functionalities which makes the \system{} extensible. \\

\begin{figure}[t]
\centering
\includegraphics[height=4.5cm, width=9cm]{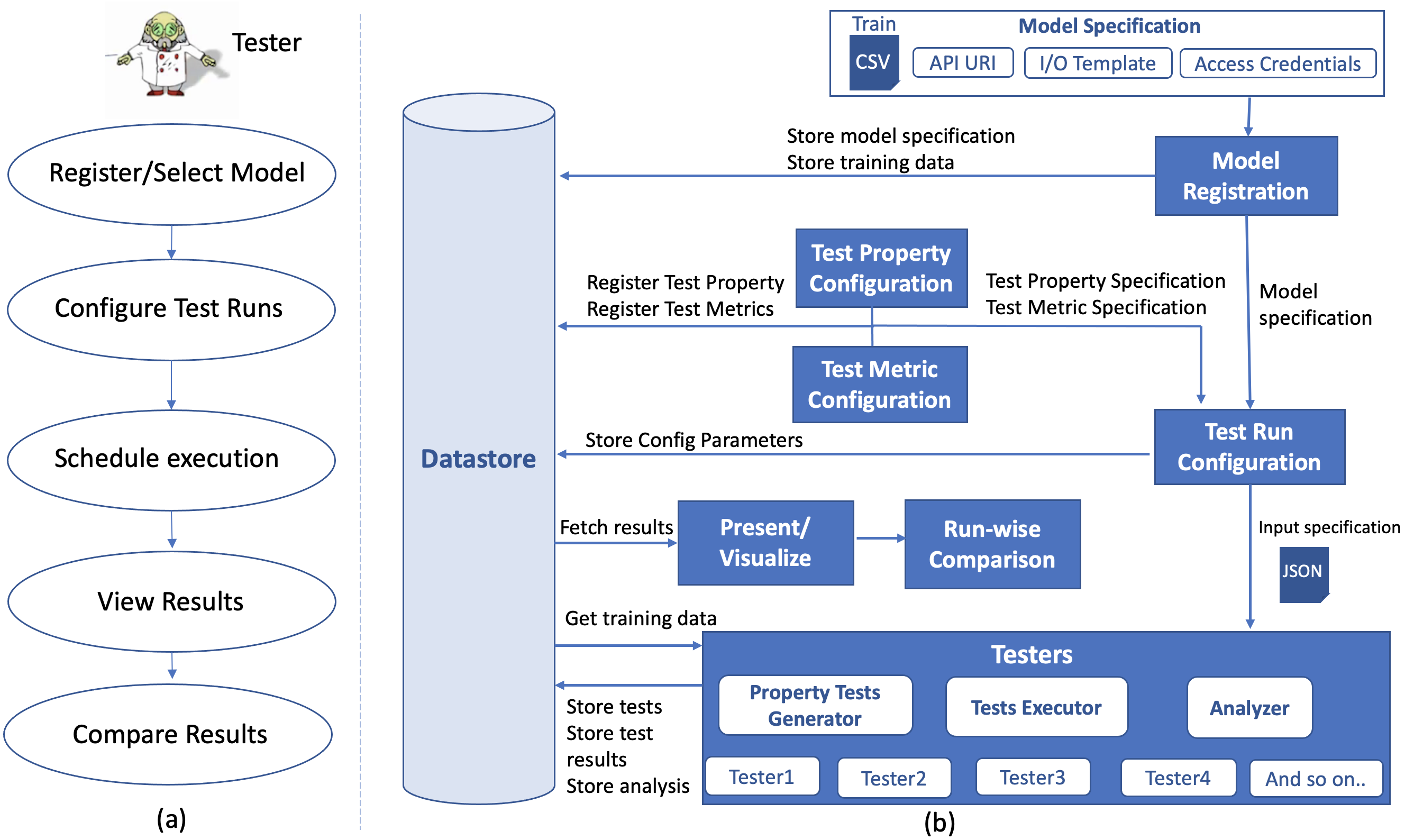}
\caption{System Flow (a) and Architecture (b)}
\label{fig:arch}
\end{figure}

\noindent{\bf Pipeline.} The pipeline is shown in Figure~\ref{fig:arch}(a). To use \system{}, an AI tester requires 1) an AI model to be hosted as an API which can output the prediction(s) corresponding to one or more sample input, 2) training data which was used to build the model. User first selects the type of model (\system{} supports tabular-data classifier, time-series prediction, text-classifier, conversation-classifier models). Then, the user registers a new model as the part of a project into \system{} by providing model API specification and uploading training data. Alternatively, the user can select an already registered model. In the next step, the user selects the model type-specific properties and necessary inputs (described later) required for each property and some data properties which are going to be considered for multiple test-properties. Users can view the status of the test run which shows the number of test cases generated and their status specific to each property. Once the test run is completed, the user can see all the metrics with some recommended action and subsequently view the failure samples along with some explanation. Users can compare the result of multiple runs for all runs in the project. Note that, the pipeline is the same for all types of models and properties. \\

\begin{figure}[b]
\centering
\includegraphics[height=4.5cm]{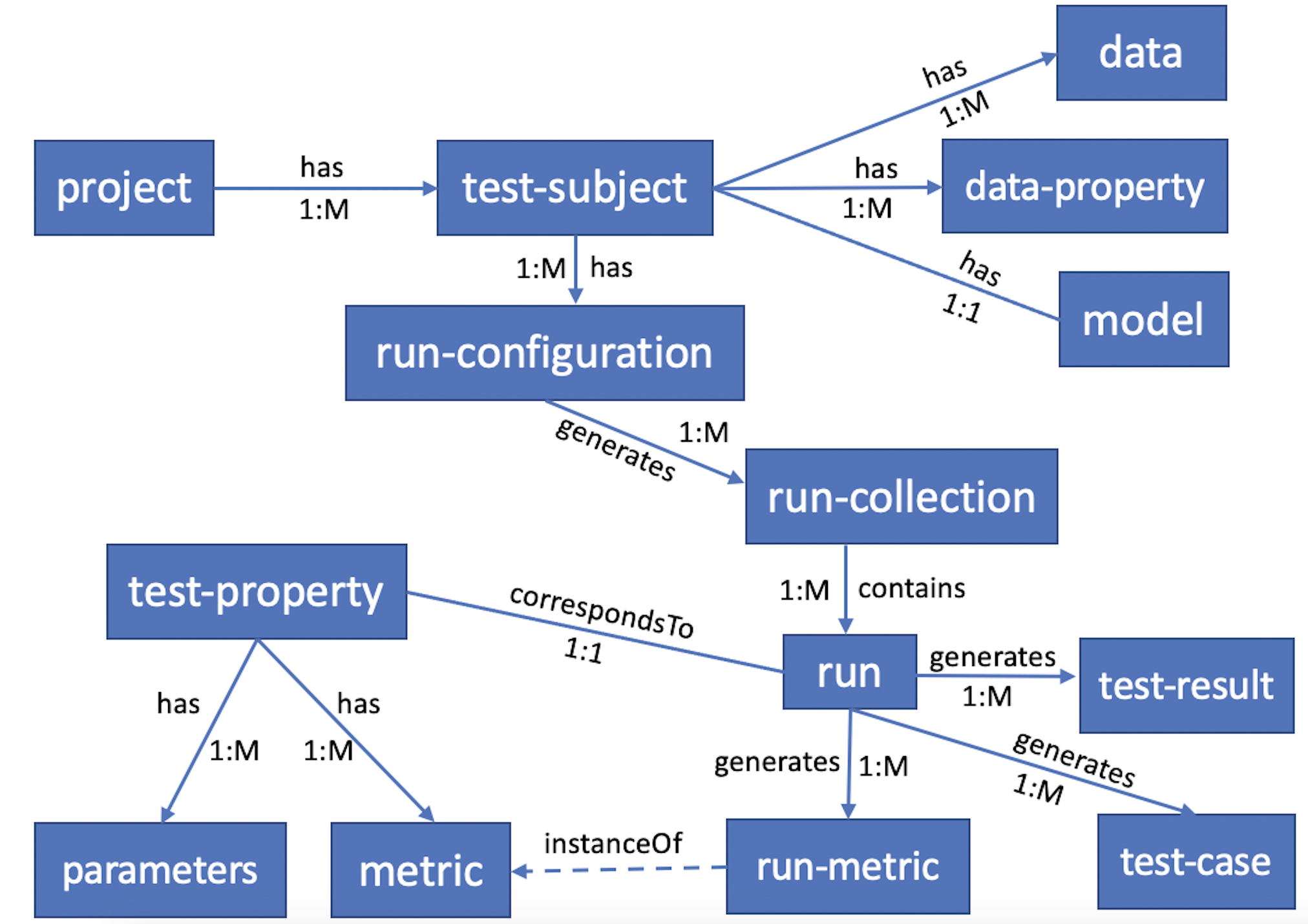}
\caption{Data Model}
\label{fig:data}
\end{figure}

\noindent{\bf Data model.} Here are some key data model design considerations. Consider three properties - correctness, individual discrimination~\cite{FTA}, and group discrimination~\cite{AIF360}. Correctness checking involves checking  a sample's prediction against the gold standard. Individual discrimination testing tries to match the predictions of two samples differing only in user-specified fairness attributes (like race, gender, etc.). Finally, group discrimination involves checking whether two groups, defined by fairness attributes, are getting sufficiently similar decisions across all the samples. As evident from these varied checks, the notion of test-case cannot be defined at a sample level (as done for software testing). We, therefore, define \emph{test-case} as a set of samples along with required reference values that can be evaluated against the model to determine success or failure. A test result is associated with a test case that contains the model predictions along with a boolean value denoting success/failure of the test case. A metric is an expression computed over \emph{all} test-cases and results. A test-case for correctness contains one sample with gold standard prediction, while for individual and group discrimination, it contains a pair and a set of samples, respectively.

This abstraction also helps to cope with the changes. Say, even if the model changes behind the model API, the system can still evaluate the existing test-cases to generate new test-results and metrics. \system{} can evaluate a newly added metric on top of existing test-cases and results.   

The data model is shown in Figure~\ref{fig:data}. One \texttt{project} can have multiple \texttt{test-subjects}. Each \texttt{test-subject} can have one \texttt{model}, multiple \texttt{data} (either training data or result-visualization data), and multiple \texttt{data-properties} (e.g. list of column names). Each \texttt{test-property} (e.g. correctness) can have multiple \texttt{metrics} (e.g. precision, recall, F-score for correctness) and multiple \texttt{parameters} (e.g. fairness threshold in group-fairness).  Multiple \texttt{run-configurations} can be defined for each \texttt{test-subject} where each \texttt{run-configuration} contains a set of \texttt{properties} (like correctness and group fairness), properties-specific parameter values (e.g., for group fairness, a range $R$ is specified with default values [0.8, 1.25]), and data specific properties (e.g. list of fairness attributes like age, race  and any change in data-distribution for \textit{what-if} testing). Each \texttt{run-configuration} can be re-used multiple-times, where each test-run will generate one \texttt{run-collection}. One \texttt{run-collection} can have multiple \texttt{run}s, each of which  corresponds to a \texttt{test-property}. Each \texttt{run} generates one value for every \texttt{metric} associated with the property (called \texttt{run-metric}) and generates a set of \texttt{test-cases} with \texttt{test-results}. 

Note that this abstraction enables us to extend the functionality of the system just by updating the data values in the data-store associated with this data model. For instance, it is possible to add a new property and its corresponding tester  by updating the data-store with relevant APIs without any changes to the code. We initialize \system{} in the same manner.

\noindent{\bf Functionalities.} In this section, we describe the functionalities of various components of \system{}. The system architecture is shown in Fig.~\ref{fig:arch}(b).  We logically divide the tool into two parts - the \texttt{framework} and the \texttt{testers}. The framework contains a UI, data-store, and interacts with the user-input, model API and testers. The testers are responsible to generate realistic test cases, obtain the prediction of these test cases using framework, compute the property metrics, and finally update the framework with metrics, test-cases and results. We describe the framework components in this section, and the functionalities of the tester components in the next one.   
\eat{

\subsection{System Components}
\subsubsection{Framework engine}
This is the main component that provides the common services such as tester registration, model registration, definition of test configurations, run configurations and presentation of the results returned by the various testers. It also has a data store at the back-end, where details of the models, tests, runs and output are persisted. The framework comprises:
}
\begin{itemize}[leftmargin=*]
\item{Model Registration:} As mentioned earlier, the framework is model-agnostic and views each model as a black-box, the framework exposes a functionality to specify model input as a template where placeholders can be replaced by the generated test data to form the legitimate call to model APIs. To retrieve the label and confidence from the model API response, we use the JSONPath patterns~\cite{JP}. Using this generic way to create API input and retrieve API output enables \system{} to connect to commercial platforms such as Watson ML, Amazon SageMaker or any custom model API very easily. User can specify additional header information, such as authentication, required to access the API. Note that individual testers do not get access to this confidential information as the framework calls the model API and testers can invoke this functionality through API, thereby \emph{preserving privacy} as we envision that testers can be developed and hosted by any third party. 

\eat{
specified only by its input and output parameters. Each of the testers defines the specifications for its models. The framework provides the support for registering the various kinds of models, for the different modalities. The details needed to specify each model are dependent on the modality of the tester, such as text, tabular data and time series. The common details include the model API, the input and output templates for the model, the details of the training data, and optional details may include access credentials. Once the details are specified, the model is registered and made available to the user for configuring and executing different kinds of tests.}
\item{Test Run Configuration:} A test run may be configured for any of the currently registered models. Different modalities support different test properties (see Section~\ref{sec:algo}) and when user selects a set of  properties for a run configuration, each tester for a selected property is invoked in a separate thread. Each tester is passed the information regarding the test-subject and the run-configuration. 
\eat{
at the time of configuring a specific test, the user may select all or a subset of these properties for testing the performance and soundness of the model against these properties. Further, based on the modality, different configuration parameters are also taken as an input. For example, for tabular modality, parameters like protected attribute, favourable label, majority protected group specified as an expression (all needed by fairness testers), additional overriding constraints as JSON formatted user-defined constraints are to be taken as an input from the user during run configuration.
All these configuration details are well persisted, so that the tests may be rerun later and results compared, as the model may evolve over multiple training sessions.  
}
\item{Results:}  The presentation component displays the status of the runs containing the total number of test generated, executed, passed, and failed. Subsequent drill down shows the metrics, failure test cases and the results. Since, test-cases and results contain different information depending on the property, the UI \emph{dynamically adapts} for every property based on the specification of the property object.

\eat{
\item{Datastore:} The framework includes a data store for persistence of the information on the various configurations at all stages. Under each of the supported modalities, the framework saves details of the models registered, the test runs configured, the tests run, and the detailed results. The data model is as shown in Figure~\ref{fig-datamodel} Currently, our datastore is implemented in Postgres, but this is lightly coupled and may be substituted by any other relational or non-relational database.  
}
\end{itemize}

\eat{
\subsection{System Architecture}
The system architecture as shown in Fig.~\ref{fig:arch} is flexible enough and supports the following features.
\begin{itemize}
\item It is designed as a plug-and-play system, where different testers may be plugged in for testing different properties for varied modalities of the input data. 
\item A test case is the building block of our system. Each test case is associated with a single test property. The user can select multiple test properties in one run, and each test property can generate and execute multiple test cases. 
\item Execution of each test property is characterised by multiple metrics which are then reported on the framework.
\item The system is model-agnostic, and each model is a black-box, specified only by its input and output parameters. Specification of these input and output templates for a model is a one-time operation which is done by the user while registering it with a particular tester.
\item Each tester may support different test properties, different test configurations and performance metrics. The underlying contract with the framework is specified as simple JSON objects.
\item  We provide a dynamic User Interface that can be used to configure a new model, orchestrate the test runs by selecting appropriate test properties and parameters, visualizing comparing the results of multiple runs from the various models. 
\end{itemize}
}

\eat{
\subsection{Workflow}
With the above components and with reference to Fig.~\ref{fig:arch} the development and runtime work flows are as follows.
\begin{enumerate}
\item The developer of a tester module registers it with the framework by specifying the test configuration parameters, test metrics and the test properties.
\item At runtime, a user selects the test modality from the interface. The user then registers a new model (or may test an already existing one.) A new model is registered by specifying the various inputs required for that modality and tester, which include the specification of model api, training data, prediction label, input and output templates and authentication details for the model.
\item After registering the model, the user selects the test properties from the list displayed on the UI based on the modality of the associated training data. 
\item The user then schedules and executes this run configuration after specifying various configuration parameters, such as maximum number of test cases to be generated.
\item The user can then view the results for the selected test properties and even compare it with the results of already existing run configurations within the same project.
\end{enumerate}
}
\section{Testing Algorithms}
\label{sec:algo}

Different testers may be integrated with the framework dynamically by specifying the corresponding test-property, associated parameters and their input format, associated metrics, and the result format. The framework UI renders this information to obtain/display the appropriate input/output. The test properties with their parameters, metrics and test failure conditions are summarized in Table~\ref{tab:1}.\\
\begin{table}[t]
  \caption{Test Properties for different Model Types}
\centering
 \begin{tabular}{|l|l|p{0.25\linewidth}|}\hline`
Test Property & Metrics & Test Failure\\\hline
\multicolumn{3}{|c|}{Tabular data Classifier}\\\hline
Correctness & Accuracy & \multirow{4}{=}{Label mismatch with gold standard label}\\
  & Precision & \\
  & Recall & \\
  & F-Score & \\\hline
Group Discrimination (R) & Disparate Impact (DI) & \multirow{2}{=}{If DI $\notin R$}\\
  & Demographic Parity & \\\hline
Individual Discrimination & \multirow{2}{*}{Flip-rate} & \multirow{2}{=}{Label mismatch of the two samples}\\\cline{1-1}
Adversarial Robustness &  & \\\hline
\multicolumn{3}{|c|}{Text Classifier}\\\hline
Typo Sensitivity(L) & \multirow{6}{*}{Flip-rate} & \multirow{6}{=}{Label mismatch of original and transformed sample} \\\cline{1-1}
Noise Sensitivity(L) & & \\\cline{1-1}
Adjective Sensitivity & & \\\cline{1-1}
Tense Sensitivity & & \\\cline{1-1}
Voice Sensitivity & & \\\cline{1-1}
Paraphrase variation & & \\\hline
\multicolumn{3}{|c|}{Time Series Forecasting}\\\hline
Small Linear Change ($\alpha$) &  \multirow{3}{*}{RMSE change ($\Delta R$)} & \multirow{2}{=}{If $\Delta R > \alpha$}\\\cline{1-1}
Un-ordered data ($\alpha$) & &\\\cline{1-1}\cline{3-3}
Large Linear Change ($\beta$) &  & If $\Delta R < \beta$ \\\hline
  \end{tabular}
  \label{tab:1}
\end{table}

\eat{
\begin{table*}[t]
  \centering
  \begin{tabular}{|p{0.10\linewidth}|p{0.20\linewidth}|p{0.20\linewidth}|p{0.20\linewidth}|p{0.20\linewidth}|}
    \hline
Modality & Test Property & Metric & Failure case & Action \\
\hline
\hline
Text & Robustness of model to various perturbations to input text. Perturbations include:
\begin{itemize}[leftmargin=*]
\item Addition of typos
\item Addition of noise
\item Variations in tense, voice, adjective intensity
\item Adversarial variations using back-translation
\end{itemize}
&
Flip rate of the class label
&
Class label should not change for perturbations –the label before and after perturbation should be the same.
&
Based on \%age of flips of class labels, retraining with additional data is recommended.\\
\hline
Conversation &
Static analysis to check for possible errors  in the chatbot attributes –entities,  intents and dialog nodes. &
Fractions of:
\begin{itemize}[leftmargin=*]
\item Intents with less examples
\item Intents with no verbs
\item Short intents
\item Intents with invalid relations
\item Dialog nodes that can pre-emptively end a conversation
\end{itemize} &

If any fraction is above a user-defined threshold, that test case is a failure. &
Report the findings for relevant action.
\\
\hline
\multirow{4}{\linewidth}{Tabular data} & Individual Fairness & Flip rate of the class label (with change in protected attribute values) & For any sample input, class label should not change with change in value of protected attribute – the label before and after perturbation should be the same. & \\ \cline{2-5}

&Group Fairness & 
\begin{itemize}[leftmargin=*]
\item Disparate impact
\item Demographic parity
\end{itemize} & If the value of disparate impact is less than the user-defined threshold (say 0.8), it is a failure & \\ \cline{2-5}
& Robustness & Flip rate of the class label (for synthetic neighboring samples) & For any sample, the class label should not change for neighboring perturbations – the label before and after perturbation should be the same. & \\ \cline{2-5} 
& Prediction Accuracy & Accuracy Score & For any sample, if its model prediction differs from the ground truth, then it's a failure & \\ \hline
\multirow{4}{\linewidth}{Time series} & Small (adversarial) linear change & RMSE gain & If RMSE gain is more than 10\%, it's a failure & Retrain with more regularization \\ \cline{2-5}
& Large Linear Change (Adding 10*(max-min) to each instance) & If RMSE gain is less than less than 10\%, it's a failure & Do not normalize with the min-max of the test data. \\ \cline{2-5}
& Zero-ranged data & Status code of API call & If the API call fails with any status code apart from 200, it's a failure & Do not normalize with the min-max of the test data. \\ \cline{2-5}
& Un-ordered data & RMSE gain & If RMSE gain is greater than 0, it's a failure & Sort the data with timestamp before attempting to forecast.\\
\hline
  \end{tabular}
  \caption{Various test properties and corresponding metrics associated with all the supported modalities}
  \label{tab:1}
\end{table*}
}
\eat{
. Each tester has a contract with the framework in terms of the test metrics, tester input requirements (defined by the model) and tester output formats that need to be processed by the framework for presentation on the UI. These are specified in various JSON formats. Each tester has a generator-executor component, an executor component and an analyser component. Each tester interacts with the framework by exposing, at run-time, an API that may be invoked by the framework. After a test has been configured, when it is executed, the framework invokes the relevant API, also sending the test data and the various test configuration properties for that test, as part of the request. For instance, for text, this may include the number of test-cases to be generated and executed by the tester, along with the test data. The tester executes the tests and returns the test results as response. Next, we mention the various testers for different modalities currently included in our framework.\\}

\eat{
\item{Text - Conversation:} Testing in Conversation modality includes analyzers for some framework (\textit{Watson Assistant}) specific characteristics in addition to the testers already discussed in Text modality. The testers evaluates the classifier that is inherent to conversation systems, while the analyzers looks through the dialog tree and the text used to train the chatbot and performs checks for any possible problems; these checks includes having very less or no examples for a dialog class ({\color{red} intents referred to as dialog class?}) in the training text, dialog class which doesn't have a verb in it's name or is too short, invalid relation between classes in the dialog tree, nodes in the dialog tree which are likely to preemptively end a conversation, asking for user input without showing acceptable answer formats. If any of these checks fail, the user is notified who can then take a mitigative action.
}

\noindent{\bf Tabular.} The testers for different properties generate synthetic data to test the model.  The main challenge is to generate realistic data (which follows the same distribution as in the training data). If the user specifies changes in the data distribution in the form of user-defined-constraints (UDC), the synthesis procedure gives more priority to the distribution specified in the UDCs and follows the distribution in training data for the other attributes. Additionally, we devise a coverage criterion called path-coverage, which enables data generation equally among various decision regions. Essentially, \system{} generates a surrogate decision tree model which imitates the behavior of the original model with high fidelity. The realistic test-cases are equally generated in the regions specified by constraints in each path of the decision tree. To generate the realistic data, we infer the marginal and joint distribution constraints underlying the data, update them with UDC (if present), and then efficiently sample from the joint distribution to generate data. The details of the properties and algorithms related to synthetic data are available in~\cite{TR}.\\
\eat{
For any black-box model trained on a tabular data, the system currently assesses its accuracy, fairness, and robustness while considering all the input configuration parameters such as user-defined constraints (UDCs) and other fairness related parameters. All the testers generate synthetic \textit{realistic} tests which adhere to the distributions present in the input model's training data as well as satisfy the UDCs, if available. The two different variants of \emph{fairness testers} generate required number of test cases specifically tailored to check for the presence of individual and group fairness. We use flip-rate as the metric for individual fairness checking where for each sample, we change the predefined set of protected attribute (like race/gender which is received as a user-input from run configuration) to create pair of test cases and check for the label mismatch against the model. Alternatively, for group fairness, we report the values of metrics like disparate impact and demographic parity based on input majority protected group and favourable prediction. The \emph{robustness tester} generates the required number of realistic test cases and checks if the model is robust. A test sample is robust if it's prediction is exactly the same as that of all its neighbouring samples. We report the measure of the percentage of non-robust samples as flip-rate, which clearly indicates the model stability and robustness. The \emph{prediction accuracy tester} however, reports the prediction accuracy on the synthetically generated tests by comparing their model predictions against the true labels. True Label assignment ideally should be a done by manual intervention, but, we instead use a \emph{Multi-model Label Generator} to guess labels for these inputs, where we train multiple classifiers on the input training data and choose the best label for test data based on the prediction and confidence of all the classifiers.\\
}

\noindent{\bf Text.} Each test property related to text classifier checks if a particular kind of text transformation to an input text results in changing the label. Typically, the transformation should preserve the meaning and thus, the label flip is a failure. The research challenge is to ensure that the text transformation results in a meaningful sentence. \system{} gives the flexibility to input the level of changes for typo and noise insertion (with parameter $L$). The adjective, tense, and voice variation involves parsing a sentence from the training data with Stanford Dependency Parser, identifying the relevant words to change, transform those words, and finally checking the result against a language model~\cite{devlin2019bert} to assert the semantic validity of the sentence. For the paraphrase generation, we use a back-translation method that translates the training sentence to 20 different languages and then, translate back to English. The results undergo a pruning that checks the preservation of nouns and verbs (in original or in the form of synonyms) to reduce non-meaningful sentences generated due to back-translation. We also test conversation models (Watson Assistant) where we also check the quality of intent, entity, and dialog flow. The details are omitted for brevity. \\

\eat{
The different testers as part of the Text modality include testers that generate different test-cases based on text variations on the input text. These variations include addition of typographical errors to the initial input, addition of noise, modifying the inputs based on adjective, tense and noise variations. In each case, the test is executed based on the modified input, and the predicted output label is compared to the label of the initial text. The test metric is a measure of the number of labels flipped, between the initial and modified output, which is an indicator of the model stability and robustness to different kinds of inputs. The results provide the number of test cases generated, executed and passed, and for the cases that do not pass, the user can drill into the details and view the initial class and the class predicted for the modified text.}

\noindent{\bf Time series.} \eat{The time series data is often treated as a sub-category of the structured data with one of the columns being a timestamp associated with the record. Since the timestamp from such data sets can be leveraged to perform many interesting tasks like forecasting and sequence modeling, it is worthwhile to devote a dedicated sub-type for time-series based models.} Generally, it is a real challenge to generate realistic data estimates for future to test the time series forecasting, classification and sequence modeling systems. Hence, metamorphic properties \cite{dwarakanath2019metamorphic} need to be developed which do not require any oracle. Similar to other modalities, the key idea behind such metamorphic properties is to see the effect of various transformations of the input data on the final output of the model. Our testers for the forecasting models are focused on the following metamorphic properties: 
\begin{itemize}[leftmargin=*]
\item \textit{Small (adversarial) linear change:} When a very small constant ($mean\_first\_order\_difference/100$) is added to each record, the RMSE of the forecast should not increase by more than a user-configurable threshold ($\alpha$).
\eat{
\item \textit{Zero ranged data:} This test quickly detects the test data normalization. With zero range, if at all there is any attempt of normalization, we might see an error. Ideally, the model should exhibit the normal workflow even without a zero-ranged data.
}
\item \textit{Un-ordered data:} Since a timestamp exists in every test data record, as long as the test data represent a particular time range, the ordering of the records should not matter. If the output changes beyond a threshold $\alpha$ by changing the ordering of the records, then it is a failure.

\item \textit{Large linear change:}
By taking the test data into the range far away from the original training data (by adding $10*(max_{training} - min_{training})$ to each record), we expect the model to show large increase in the error. If the error is less than the threshold $\beta$, it indicates that the model normalizes the input test data which is incorrect. \\
\end{itemize}

\eat{
Most other properties from \cite{dwarakanath2019metamorphic} expect white-box access to the model which is not possible to generalize across various kinds of models. The framework needs only black-box access to the model, which makes it general enough to test any model that consumes the input in any of the supported modalities.}

\noindent{\bf Evaluation.} The text-classifier testers have been used to test 6 client models and in every case, it found issues in the model. The failure test cases were used to retrain the model which then increased the accuracy and robustness of the models. We do not present the details of the models and issues found due to client confidentiality. The testers for classifiers related to tabular data have been extensively tested with models built on open-source data~\cite{TR}. On average, \system{} generated test cases covers 50\% more paths than random data. \system{}-generated tests fail individual discrimination and robustness testing on an average $\approx$$16\%$ and $\approx$$45\%$, resp., more decision paths than the test-split. Synthesis using UDC generates varied test suites which presents significantly different (15\% to 194\% across properties) test results when compared to test data generated from train-test split.   
\section{Related Work}
\label{sec:related}

To the best of our knowledge, this is the first tool related to AI testing which covers such a comprehensive set of properties for different kinds of models. Concurrently, Ribeiro et al. created an open-source tool called CHECKLIST which tests various properties of NLP models~\cite{CHECKLIST}. There exist multiple techniques for effective testing of individual, group fairness and robustness for tabular data for which we have leveraged our work~\cite{TR,testing-fse, AIF360} towards this. To the best of our knowledge, our techniques for black-box testing for time-series forecasting models is a first of a kind. However, we are influenced by the metamorphic properties used in white-box testing of forecasting models in \cite{dwarakanath2019metamorphic}. 
\section{Conclusion and Future Work}
\label{sec:conclusion}

In this paper, we presented an AI Testing Framework that enables the users to perform automated testing of the black-box AI models by synthetic generation of realistic test cases. The framework can test any target model as long as its input modalities are supported. As of now, our implementation supports models trained with tabular, text(plain), text(conversation) and time-series data, but the design is generic enough to accommodate many more modalities. With this framework, we enable the AI testers and developers to test their models effectively irrespective of the lack of a sufficient amount of realistic labeled test data and \emph{what-if} scenarios.

In future, we plan to add support for testing models on images, videos, audio, and multi-modal inputs. The current implementation only supports black-box testing, which when configured, can be applied to a large number of other similar models of the same type without much change. We will also add support for white-box automated testing of some popular and frequently used models.

Our experimental open-source datasets are available at \url{https://github.com/aniyaagg/AI-Testing}, but we are not able to disclose rest of the artifacts because of confidentiality clause.

\bibliographystyle{IEEEtran}
\bibliography{IEEEabrv,bib}

\end{document}